\documentclass[conference]{IEEEtran}
\IEEEoverridecommandlockouts
\usepackage{cite}
\usepackage{amsmath,amssymb,amsfonts}
\usepackage{algorithmic}
\usepackage{graphicx}
\usepackage{xcolor}
\def\BibTeX{{\rm B\kern-.05em{\sc i\kern-.025em b}\kern-.08em
    T\kern-.1667em\lower.7ex\hbox{E}\kern-.125emX}}
\begin{document}

\title{Learning of Robot Safety Policies via Adversarial Synthetic Scenarios}

\author{
\IEEEauthorblockN{Nikolai Dorofeev}
\IEEEauthorblockA{SafePI.ai\\
Grenoble, France\\
contact@d0rich.me}
\and
\IEEEauthorblockN{Alexey Odinokov}
\IEEEauthorblockA{SafePI.ai\\
Belgrade, Serbia\\
ao@xortech.rs}
\and
\IEEEauthorblockN{Rostislav Yavorskiy}
\IEEEauthorblockA{SafePI.ai\\
Madrid, Spain \\
ryavorsky@gmail.com}
}

\maketitle

\begin{abstract}
Ensuring the safety of robotic systems operating in open-ended, human-centric environments remains a fundamental challenge. While recent advances in foundation models enable robots to generalize across tasks via in-context learning and fine-tuning, these approaches lack systematic mechanisms for incorporating structured safety knowledge. At the same time, hazard-informed engineering pipelines provide a principled framework for modeling risk, but suffer from a combinatorial explosion of possible scenarios when applied at scale.

In this work, we propose an agentic gamification framework for hazard-informed learning of robot safety policies through synthetic scenarios. We model scenario generation as an adversarial game between two agents: a Red Team that explores the space of potential failures by constructing hazardous situations, and a Blue Team that incrementally refines safety policies to prevent them. This iterative process enables efficient discovery of high-risk edge cases that are unlikely to be captured through random simulation or manual enumeration.

By combining classical risk modeling with adversarial scenario generation and modern learning paradigms, this work provides a scalable pathway for embedding safety into Physical AI systems operating in complex real-world environments.
\end{abstract}


\section{Introduction}

The increasing deployment of robotic systems in unstructured human environments, such as homes, hospitals, and public spaces etc., poses a fundamental challenge: how to ensure safety under both predictable failures and open-ended, context-dependent interactions. Unlike traditional industrial robotics, where operating conditions are tightly controlled, modern Physical AI systems must function in dynamic settings characterized by uncertainty, variability, and continuous human presence.

Recent advances in large-scale machine learning, particularly in-context learning and fine-tuning of foundation models, have significantly expanded the capabilities of robotic systems. Robots can now generalize to novel tasks, interpret natural language instructions, and adapt behavior without explicit reprogramming. However, this flexibility introduces a critical safety gap: models trained primarily for task performance may lack systematic grounding in hazard awareness. As a result, they can produce actions that are functionally correct yet physically unsafe.

To address this, we build on the concept of a hazard-informed engineering pipeline, in which safety is not treated as an external constraint but as a first-class design principle embedded throughout the learning process. As outlined in the subsequent section, this pipeline structures safety engineering into five stages: asset declaration, exposure modeling, hazard scenario definition, simulation-based data generation, and safety-oriented model training. This approach enables the systematic translation of informal safety knowledge into formal representations that can be used to generate training data and guide model behavior.

However, a key limitation emerges when attempting to operationalize this pipeline at scale. Even in relatively simple environments, the number of possible interactions between objects, robot actions, environmental conditions, and human behaviors grows combinatorially. Exhaustive enumeration of all hazard scenarios becomes infeasible, particularly when considering rare but safety-critical edge cases that arise from complex system interactions.

This paper addresses this scalability challenge by introducing an agentic gamification framework for scenario generation. Instead of attempting to enumerate the full hazard space, we reformulate the problem as an adversarial exploration process, where scenarios are iteratively constructed to expose weaknesses in the current safety policy. This approach enables efficient discovery of high-value edge cases and produces structured synthetic datasets tailored for safety-critical fine-tuning.

By integrating hazard-informed reasoning with adversarial scenario generation and modern learning paradigms, we aim to bridge the gap between classical safety engineering and data-driven robotics, enabling systems that are not only capable, but systematically aligned with safety constraints in real-world environments

This paper describes ongoing work. We present the conceptual framework, a small-scale proof-of-concept experiment, and a roadmap for future validation. The contribution is a problem formulation and a proposed solution architecture, rather than a fully evaluated system.

\subsection{A Hazard-Informed Pipeline for Robotic Physical Safety}

To systematically address both deterministic and emergent harm, we propose a five-step hazard-informed pipeline that integrates classical risk analysis with modern machine learning workflows:

{\bf Step 1: Asset declaration.} 
Exhaustively enumerate all assets requiring protectio: human (operator, bystander), organizational (hardware, reputation), and environmental (soil, air), at multiple granularities, without filtering.

{\bf Step 2: Exposure modes.}
For each asset, identify how it could become vulnerable to harm (e.g., human limb exposed to moving actuator, battery exposed to overheating), establishing a taxonomy of weaknesses independent of specific causes.

{\bf Step 3: Hazard scenario definition.} 
Transform abstract vulnerabilities into concrete, testable scenarios by mapping each exposure mode to a causal chain (e.g., sensor occlusion → failed human detection → collision).

{\bf Step 4: Simulated scene and synthetic data generation.}
For each hazard scenario, construct a digital twin, inject the failure mode, and generate thousands of annotated variations (lighting, occlusions, sensor noise) to create structured datasets grounded in explicit hazard modeling.

{\bf Step 5: ML fine-tuning and safety envelope learning. }
Use these datasets to fine-tune perception models, train anomaly detectors, and develop hazard anticipation models that enable the robot to recognize and avoid unsafe states before they occur.

This pipeline provides a systematic, auditable methodology for embedding safety into robotic systems across both predictable failures and complex emergent risks.

\begin{figure}[htbp]
\centerline{\includegraphics[width=0.49\textwidth]{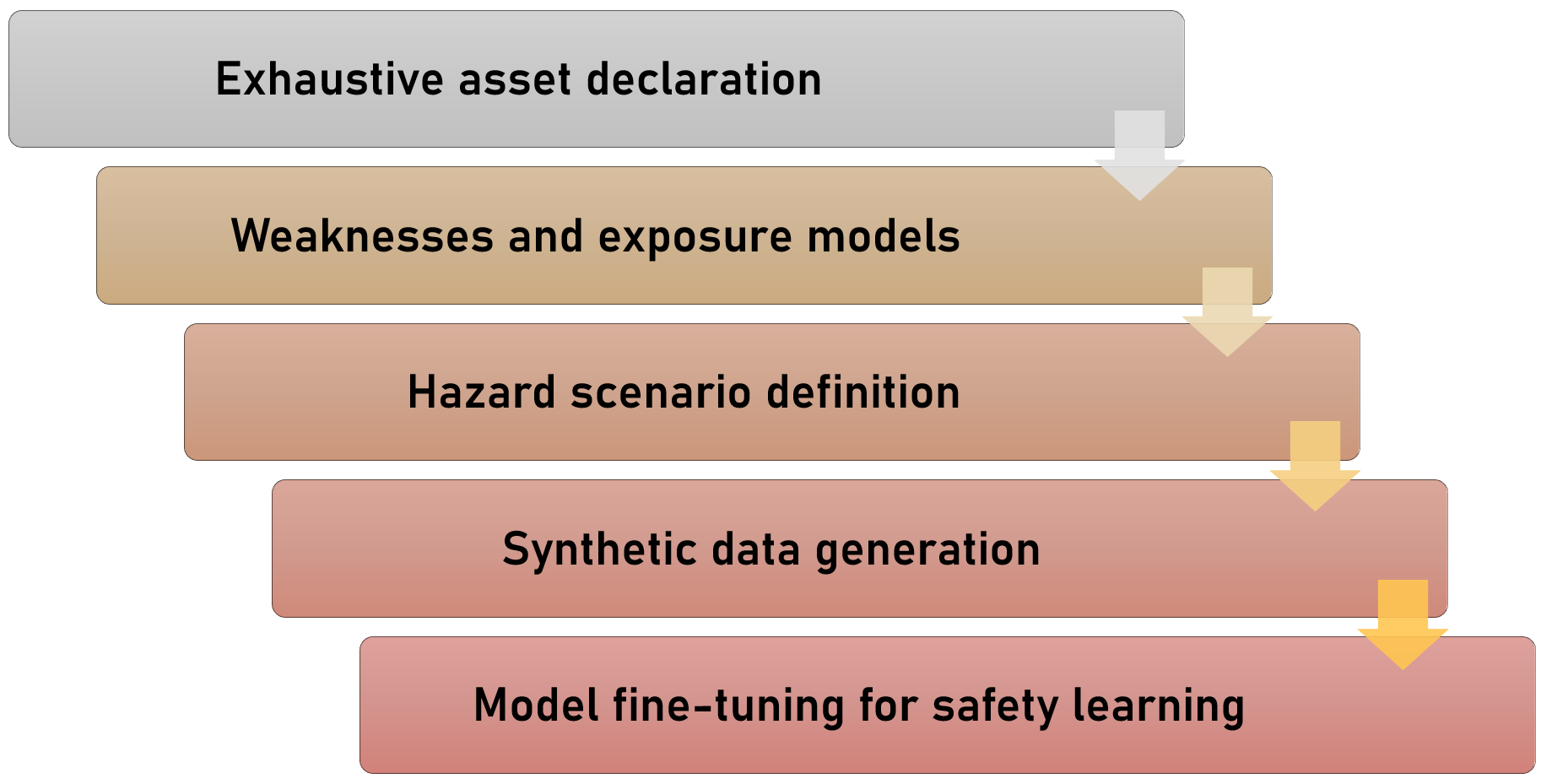}}
\caption{A Hazard-Informed Engineering Pipeline for Robotic Physical Safety.}
\label{fig_wf}
\end{figure}

\section{Formalization of the Hazard-Informed Pipeline}

To improve precision, in this section we formalize the core concepts underlying the proposed hazard-informed pipeline.

Let $\mathbf{S}$ denote the space of system states, and let $\mathbf{C}$ denote the set of all valid scenarios, where each scenario is a finite sequence of states connected by admissible system transitions. Formally, a scenario is defined as
\begin{equation}
c = (s_1, s_2, \ldots, s_n), \quad s_i \in \mathbf{S}.
\end{equation}
A \emph{hazard} is defined as a function over system states:
\begin{equation}
h: \mathbf{S} \rightarrow \{0,1\},
\end{equation}
where $h(s) = 1$ indicates that state $s \in \mathbf{S}$ is hazardous, and $h(s) = 0$ otherwise.
Similarly, a \emph{safety rule} is defined as a constraint over system states:
\begin{equation}
r: \mathbf{S} \rightarrow \{0,1\},
\end{equation}
where $r(s) = 1$ denotes that state $s$ satisfies the safety constraint, and $r(s) = 0$ indicates a violation.
A scenario $c = (s_1, \ldots, s_n) \in \mathbf{C}$ is called \emph{hazardous} if there exists at least one state $s_i$ such that $h(s_i) = 1.$ 
Similarly, a scenario is said to \emph{violate a safety rule} if there exists at least one state $s_i$ such that
$r(s_i) = 0.$

Let $\mathbf{A}$ denote the set of assets and $\mathbf{E}$ the set of exposure modes. An \emph{exposure mode} captures a specific way in which an asset may become vulnerable to harm.
We define the exposure mapping as:
\begin{equation}
E: \mathbf{A} \rightarrow 2^{\mathbf{E}},
\end{equation}
where $E(a) \subseteq \mathbf{E}$ denotes the set of exposure modes associated with asset $a \in \mathbf{A}$.

Furthermore, we define a mapping from exposure modes to  the subset of hazard scenarios:
\begin{equation}
G: \mathbf{E} \rightarrow 2^{\mathbf{C}},
\end{equation}
such that $G(e)$ represents the set of {\em hazardous scenarios derived from exposure mode} $e \in \mathbf{E}$.
Exposure modes determine which aspects of the system and environment should be perturbed, while the mapping $G$ identifies the corresponding classes of scenarios that must be instantiated in simulation.

In practice, this enables the construction of targeted simulation pipelines in which scenario generation is not random, but guided by explicitly defined vulnerabilities and hazard mechanisms. For each exposure mode, simulation parameters such as geometry, agent behavior, sensor conditions, and environmental factors can be systematically varied to produce diverse realizations of the associated scenarios. The hazard and safety rule functions then serve as labeling mechanisms, allowing each simulated state or scenario to be automatically annotated as safe or unsafe.

\subsection{Literature Overview}

The integration of large language models into robotics has opened new paradigms for robot learning and control. Two complementary approaches have emerged as particularly promising: in-context learning, where models adapt to new tasks through prompted examples without weight updates, and fine-tuning, where pre-trained models are adapted through continued optimization on task-specific data. This overview synthesizes recent advances acrossseveral papers that explore these techniques for robotics applications ranging from dynamic control to imitation learning and humanoid manipulation, with attention to how these methods might be evaluated through systematic hazard-informed pipelines.

The InCoRo system \cite{zhu2024incoro} represents a foundational approach to integrating LLMs with classical robotics feedback mechanisms. Unlike prior work that generates static execution plans, InCoRo implements a continuous control loop comprising an LLM controller, scene understanding unit, and robotic hardware. The system processes user text instructions through a pre-processor that decomposes complex commands into atomic actions and object lists, then maintains a feedback loop where the LLM controller receives real-time perceptual updates to adjust execution. This enables zero-shot generalization to novel tasks through in-context learning with off-the-shelf LLMs, while a pre-execution filter rejects inappropriate commands and recycles them as negative examples. Experimental validation on SCARA and DELTA robots demonstrates significant improvements over static planning approaches, with ablation studies revealing critical dependencies on specific system components.

Keypoint Action Tokens \cite{di2024keypoint} introduces a novel framework for few-shot imitation learning by repurposing text-pretrained Transformers as general sequence-to-sequence learners. The method transforms visual observations into three-dimensional keypoint tokens via vision foundation models, and represents end-effector trajectories as sequences of point triplets that uniquely encode poses. Critical insights include the importance of representation design and vision backbone selection, with certain self-supervised models proving more effective than those pretrained on robotics data. Notably, the work demonstrates that more recent language models show progressive improvement, meaning that advances in language modeling directly benefit robotics without domain-specific innovation. The method excels with very few demonstrations but plateaus with more data, suggesting fundamentally different scaling properties than fine-tuning.

The RoboMorph approach \cite{bazzi2024robomorph} applies encoder-decoder Transformers to learn meta-dynamical models without explicit physical parameters. Using massively parallel simulations, the model learns to predict end-effector poses and joint positions from torque signals across diverse dynamics configurations achieved through domain randomization. The system uses a portion of each trajectory as context to predict the remainder, demonstrating zero-shot generalization within the same task family. Importantly, pre-training on diverse signals followed by relatively brief fine-tuning on specific tasks outperforms much longer training from scratch. However, the model struggles to generalize across fundamentally different control action families, suggesting that black-box approaches cannot fully replace physics-based modeling for arbitrary dynamics.

InCLET \cite{wang2025inclet} addresses natural language-conditioned reinforcement learning by leveraging in-context learning to generate task representations. The method prompts the LLM to imagine multiple task-relevant trajectory pairs showing initial and terminal states, extracts hidden states from across these demonstrations, and projects the high-dimensional representation to a lower-dimensional space for policy conditioning. Theoretical analysis reveals the fundamental trade-off inherent in this approach: better task representations improve performance but may increase model complexity. Empirical results demonstrate that this method outperforms both direct LLM embeddings and one-hot encoding approaches, particularly on unseen natural language instructions, with visualization confirming more tightly clustered task representations.

MIMICDROID \cite{shah2025mimicdroid} addresses the scalability challenge by leveraging human play videos, which are substantially faster to collect than teleoperated demonstrations and inherently capture diverse manipulation behaviors. The method extracts trajectory pairs with similar manipulation patterns via feature similarity, then trains a transformer policy to predict actions of one trajectory conditioned on the other. Key innovations include extracting action information from RGB video via hand pose estimation, applying visual masking to reduce overfitting to human-specific appearances, and kinematically retargeting from human to humanoid poses. The work introduces a systematic simulation benchmark with three generalization levels, testing performance on seen objects, unseen objects, and entirely novel environments. Results demonstrate substantial improvement over baselines, with scaling analysis revealing consistent gains from more training data and more in-context examples up to context length limitations.

Examining these works through a hazard-informed lens reveals both strengths and gaps in current approaches. Regarding asset declaration, all methods implicitly protect the robot itself and task-relevant objects, but systematic enumeration of human operators, bystanders, and environmental assets remains absent. For exposure modes, the reviewed systems address certain vulnerabilities, InCoRo's feedback loop detects unexpected object displacements, while MIMICDROID's visual masking reduces overfitting to human appearance, yet none establish comprehensive taxonomies of weaknesses independent of specific causes.

Besides, there are quite many papers that address the critical challenge of ensuring safety in AI-driven robotic systems, spanning two complementary domains: LLM-controlled agents and contact-rich physical manipulation.

In \cite{yang2024plug} a ``safety chip'' is introduced that uses Linear Temporal Logic (LTL) to enforce user-defined safety constraints for LLM-driven robots. The main result is that their system, which translates natural language constraints into LTL for monitoring and action pruning, achieves a 100\% safety rate in experiments, significantly outperforming baseline LLMs that struggle with complex safety rules.

SafeEmbodAI \cite{zhang2024safeembodai} proposes a framework to secure LLM-controlled mobile robots against threats like malicious prompt injections. The key finding is that their method, which combines secure prompting, state management, and safety validation, leads to a substantial 267\% performance improvement in complex environments under attack, demonstrating its effectiveness in mitigating unsafe behaviors.

LongSafety \cite{lu2025longsafety} presents the first benchmark to evaluate LLM safety specifically in long-context tasks. The main result reveals a significant safety vulnerability, with most models achieving safety rates below 55\%, and shows that strong performance in short-context tasks does not guarantee safety when models are given long, complex contexts.

SELP \cite{wu2025selp} introduces an LLM planner that combines equivalence voting for translating commands into LTL, constrained decoding for safe action selection, and fine-tuning for efficiency. The results show SELP outperforms state-of-the-art planners, improving safety rates by up to 20.4\% and efficiency by 19.8\% in drone navigation and robot manipulation tasks.

The paper \cite{moeini2025safe} pioneers the study of safety in In-Context Reinforcement Learning (ICRL), where agents adapt to new tasks without parameter updates. Their proposed method, SCARED, is the first to enable safe in-context adaptation, demonstrating that it can enforce safety budgets and outperform existing safe meta-RL and algorithm distillation baselines on out-of-distribution tasks.

A survey \cite{zhang2025safe} provides a comprehensive overview of safe learning methods for robots performing tasks involving physical contact. It uniquely categorizes approaches from classical methods to emerging foundation models, highlighting that while Vision-Language-Action models offer new safety opportunities, they also introduce amplified risks due to a scarcity of real-world contact data and challenges in grounding semantic constraints.

\section{The Home Robot Safety Adaptation Problem}

Consider the deployment of a general-purpose home robot entering a new, previously unseen environment. Unlike industrial settings, where both tasks and surroundings are tightly specified, domestic environments are highly heterogeneous, continuously evolving, and deeply personalized. Each household differs not only in its physical layout, but also in the properties of objects, user preferences, and implicit safety norms.

At deployment time, the robot is not starting from scratch. It is equipped with:

\begin{itemize}
\item Generic foundation models for perception and manipulation

\item Universal safety rules (e.g., ``do not apply force exceeding the fragility threshold of a typical ceramic cup'')

\item Baseline hazard associations (e.g., ``water near electrical outlets presents a shock risk'')
\end{itemize}

While these capabilities provide a strong starting point, they are inherently context-agnostic. Safe operation, however, depends critically on the specific instantiation of the environment.

For example, a policy such as ``handle fragile objects carefully'' is insufficient without understanding what fragile means in a particular home. A ceramic cup used daily may tolerate moderate handling, whereas an antique porcelain vase may require extreme caution. Similarly, a general rule like ``do not place objects near the table edge'' must be grounded in the precise geometry of the table, the friction properties of its surface, and even the habits of the household occupants (e.g., presence of children or pets). So, safety must be adapted from generic knowledge to environment-specific constraints.

Safety requires coverage of rare, high-impact scenarios, but those scenarios are precisely the hardest to enumerate and collect. This observation directly motivates the need for scalable mechanisms to explore and generate such scenarios, which we address in the following sections through an agentic, adversarial formulation.

\section{Experimental Results}

As this work represents research in progress, we begin by describing our initial foray into safety learning for robotics, a deliberately simplified experiment that nonetheless provided critical insights and directly motivated the more ambitious gamification framework proposed in this paper.

Our starting point was a basic question: Given a well-defined safety scenario, can synthetic data from a simulator be used to teach a robot to distinguish safe from unsafe situations? 

We used the Webots simulator \cite{michel2004cyberbotics, farley2022pick} to construct a simplified indoor environment consisting of a table and a single object, a can. The task was to distinguish between safe and unsafe object placement with respect to the table edge.

Two classes of scenarios were generated:
\begin{itemize}
 
\item Safe: the can is placed at a sufficient distance from the table edge.

\item Unsafe: the can is positioned too close to the edge, representing a potential falling hazard.
\end{itemize}

Using the simulator, we generated multiple videos under varying conditions (viewpoints, lighting, minor positional variations). From these videos, frames were sampled to construct a labeled dataset for training and evaluation.

\begin{figure}[htbp]
\centerline{\includegraphics[width=0.49\textwidth]{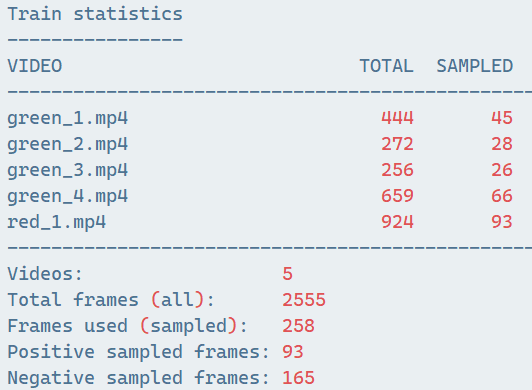}}
\caption{Train dataset statistics.}
\label{fig_t1}
\end{figure}

The dataset consists of 5 videos with both safe and unsafe configurations, see \ref{fig_t1}. 

\begin{figure}[htbp]
\centerline{\includegraphics[width=0.49\textwidth]{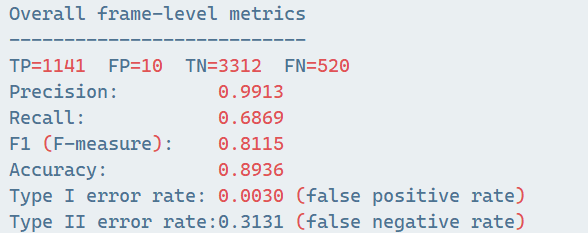}}
\caption{The trained model evaluation statistics.}
\label{fig_t2}
\end{figure}

This setup intentionally reflects a controlled but imbalanced scenario, where unsafe cases are less frequent but safety-critical. 
The trained model was evaluated at the frame level, see fig.~\ref{fig_t2}. 

The results reveal several important insights.

First, the model achieves very high precision 0.99, indicating that when it predicts an unsafe scenario, it is almost always correct. This is a desirable property in safety-critical systems, as false alarms are minimized and operator trust is preserved.

However, the recall 0.69 indicates that a non-negligible portion of unsafe scenarios is not detected. This corresponds to a relatively high Type II error rate, meaning that some hazardous configurations are classified as safe.

\begin{figure*}
\centerline{\includegraphics[width=0.44\textwidth]{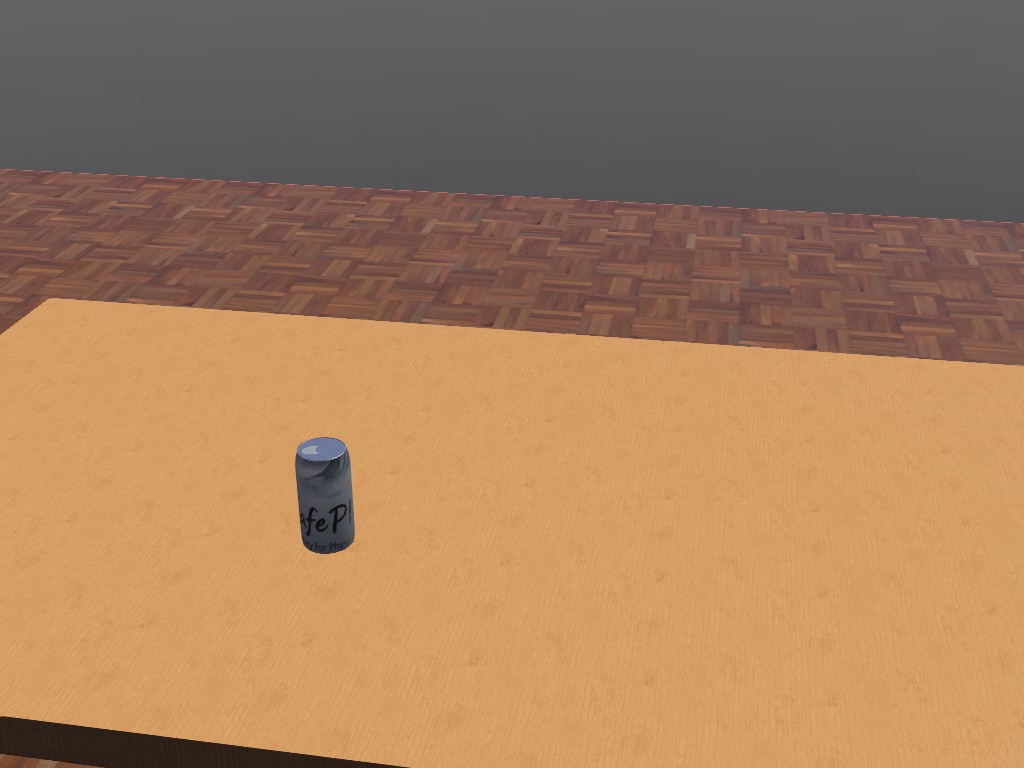}
\includegraphics[width=0.44\textwidth]{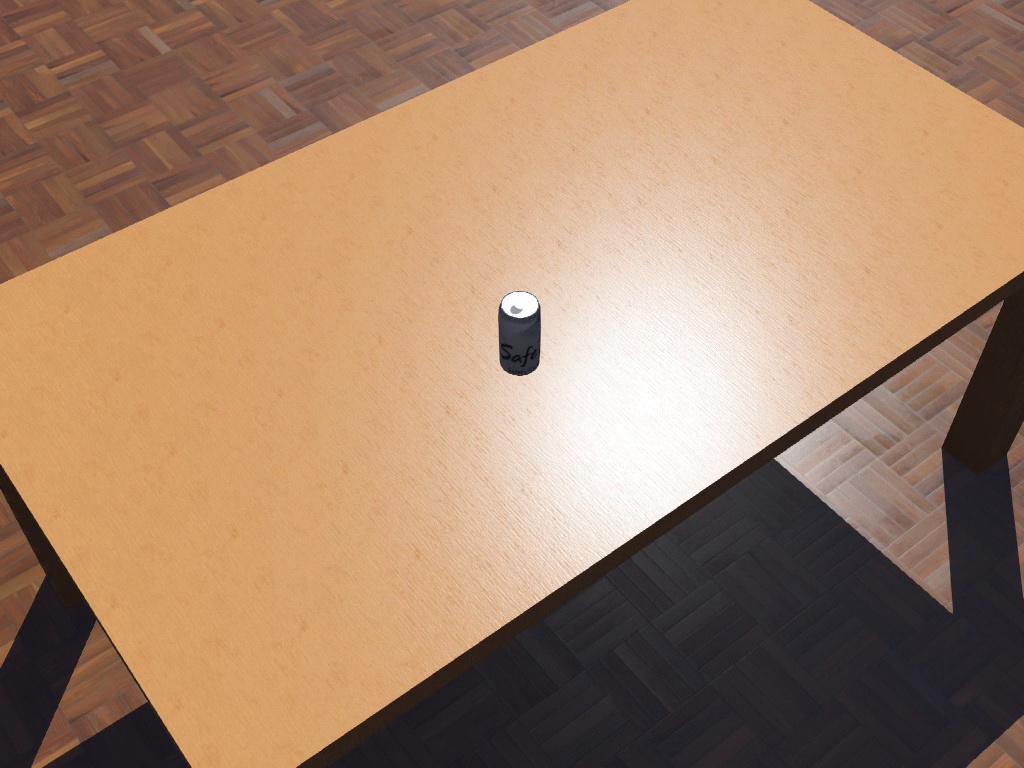}}
\caption{Images from the training dataset that correspond to a safe location of a can on a table}
\label{fig_safe}
\end{figure*}

\begin{figure*}
\centerline{\includegraphics[width=0.44\textwidth]{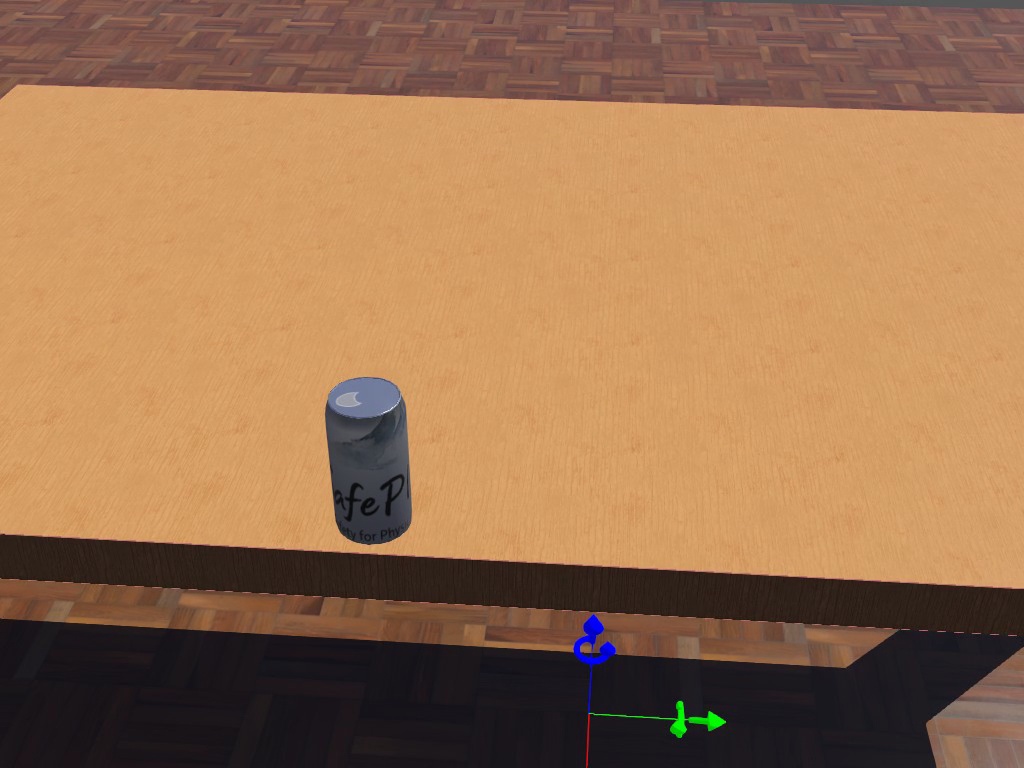}
\includegraphics[width=0.44\textwidth]{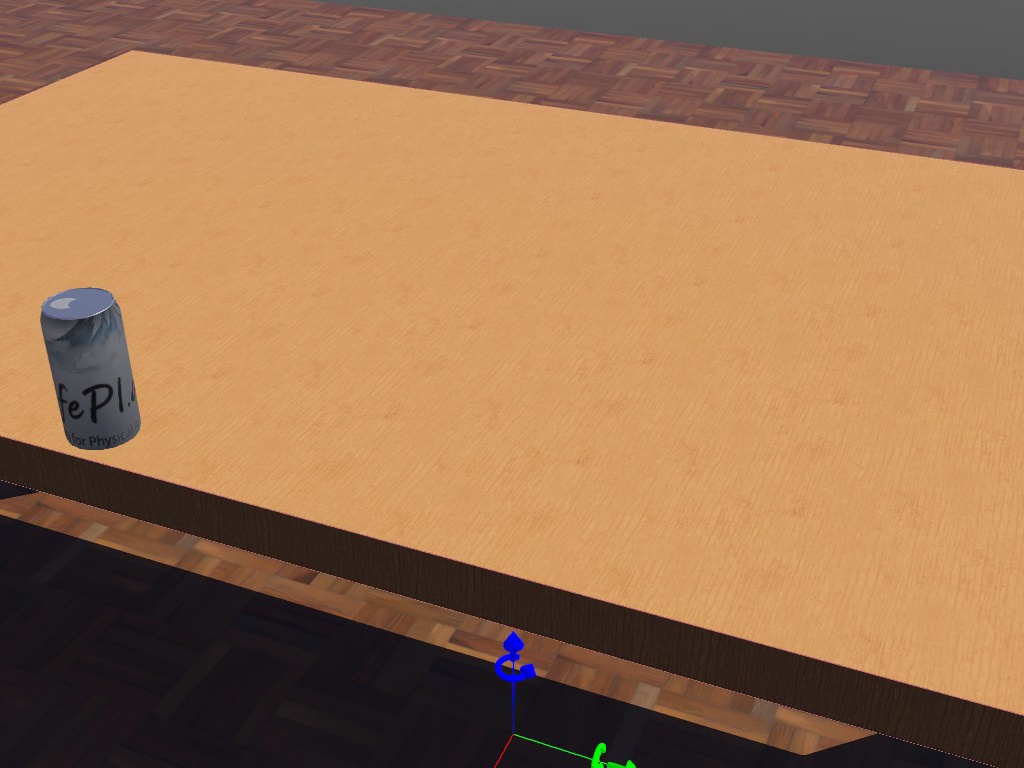}}
\caption{Images from the training dataset that correspond to an unsafe location of a can on a table}
\label{fig_nosafe}
\end{figure*}

\begin{figure*}
\centerline{\includegraphics[width=0.43\textwidth]{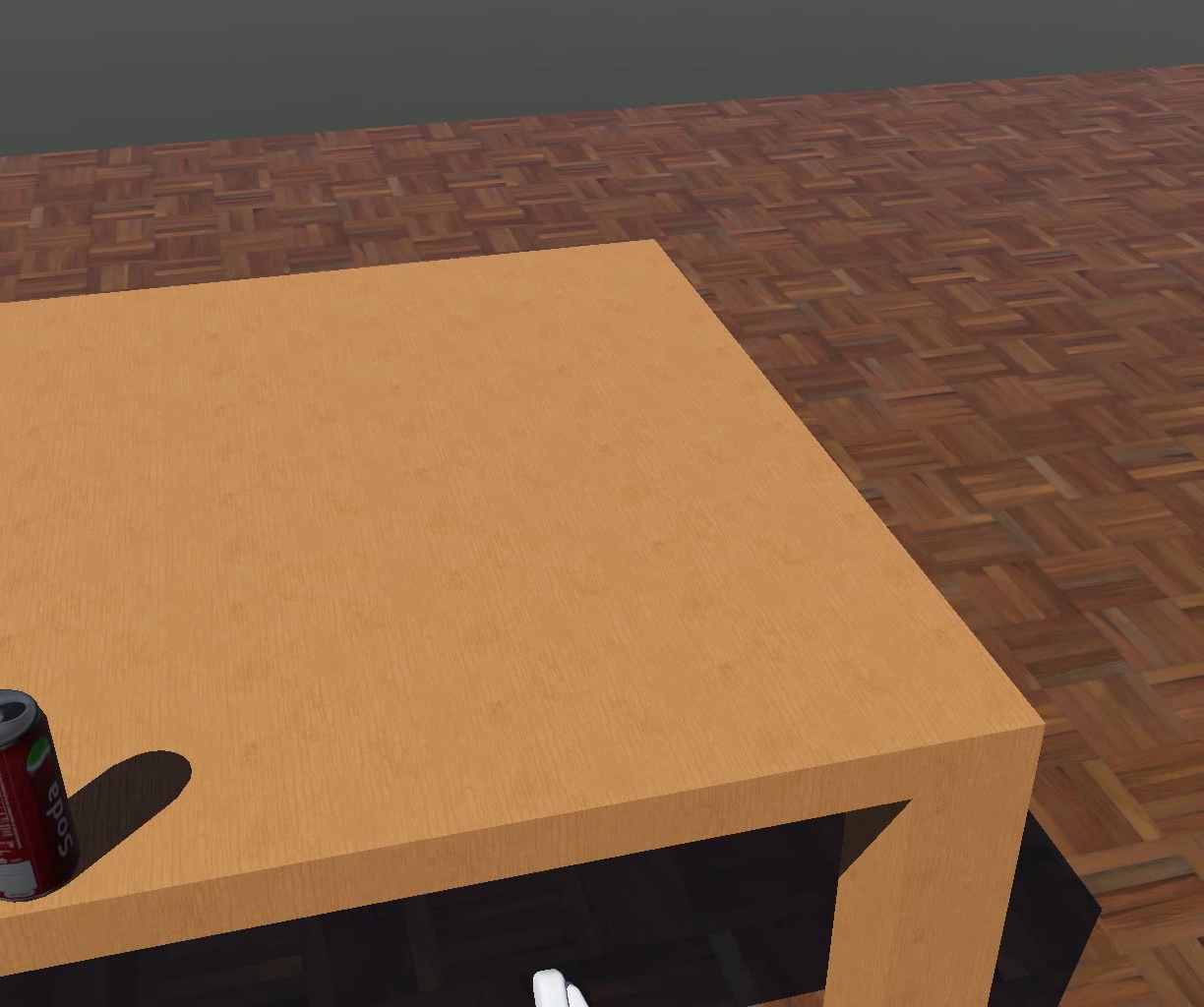}
\includegraphics[width=0.43\textwidth]{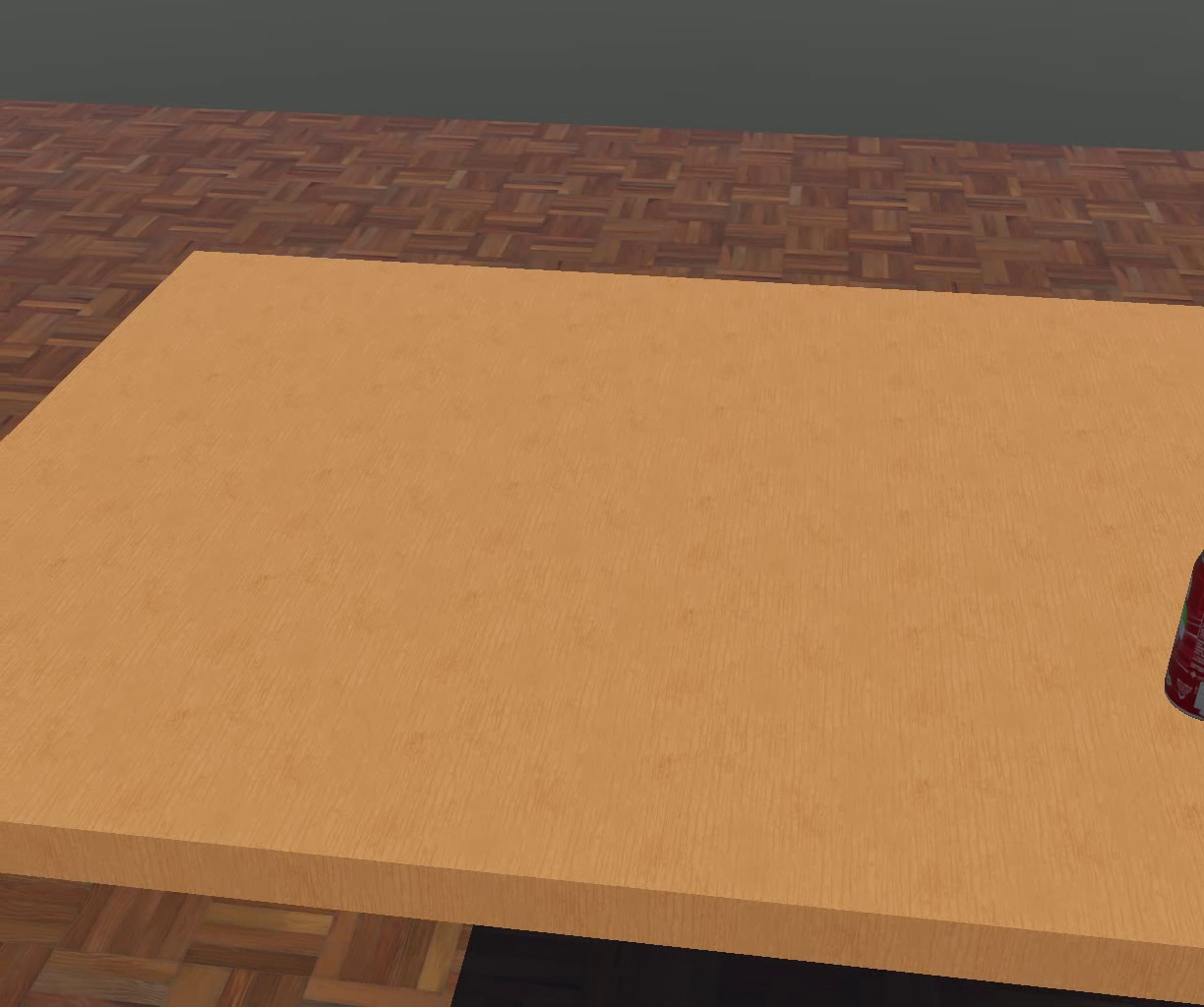}}
\caption{Images from the testing dataset when the model failed to recognize the policy}
\label{fig_fail}
\end{figure*}

This experiment, although simple, illustrates several key aspects of the proposed framework:
\begin{itemize}
\item {\em Feasibility of synthetic safety data.}
Even a small, simulator-generated dataset can train a model to distinguish safety-relevant spatial configurations.

\item {\em Importance of boundary cases.}
The most challenging examples are those near the decision boundary (e.g., marginal distances from the table edge), which are underrepresented in naive sampling.

\item {\em Need for targeted scenario generation.}
The relatively high false negative rate suggests that the dataset does not sufficiently cover hard unsafe cases, precisely the scenarios that the proposed adversarial gamification framework described below is designed to generate.

\item {\em Alignment with safety envelope learning.}
The classifier implicitly learns a boundary between safe and unsafe placements, supporting the notion of a learned safety envelope.

\end{itemize}

This preliminary experiment, despite its simplicity, clarified the fundamental challenge we face. The difficulty is not in learning a safety policy once the scenarios are defined. The core challenge lies in discovering the full set of scenarios that the safety policy must address.
And if the bottleneck is scenario discovery rather than scenario learning, then we need a mechanism that actively explores the space of possible failures. The adversarial gamification framework presented in the following sections is our answer to this challenge.

\section{The Combinatorial Explosion Challenge}

A central obstacle in operationalizing hazard-informed safety pipelines is the sheer scale of the scenario space that must be considered. Even in seemingly simple environments, the number of possible interactions between a robot, its surroundings, and humans grows combinatorially, quickly exceeding the limits of exhaustive analysis.

At a high level, each potential scenario can be understood as a combination of several factors:

\begin{itemize}

\item Objects (types, poses, states)

\item Robot actions (grasp types, trajectories, speeds)

\item Environmental conditions (lighting, layout variations, transient obstacles)

\item Human presence and behavior (poses, activities, unexpected movements)
\end{itemize}

The full scenario space is effectively the Cartesian product of these dimensions. Even under conservative assumptions --- tens of objects, a handful of action primitives, and a limited set of environmental variations --- the resulting number of possible configurations becomes astronomically large.

However, the challenge is not merely quantitative. The structure of this space is highly uneven. 
A vast majority of scenarios are benign and uninformative. 
A small fraction are safety-critical but rare.
Besides, the most dangerous cases often lie in the long tail, emerging from subtle and unlikely combinations of factors.

For example, placing a cup on a table is typically safe. Yet, under a specific combination with slightly misestimated edge position, low-friction surface, and a human bumping the table, the same action can lead to a hazardous outcome. These edge cases are precisely the scenarios that must be captured to ensure robust safety, yet they are the least likely to be discovered through random sampling or standard testing procedures.

It is clear that effective safety learning requires not uniform coverage of the scenario space, but targeted exploration of its most informative regions, particularly near the boundary between safe and unsafe behavior.

The problem is not simply one of generating more data, but of generating the right data, scenarios that are maximally informative for refining safety policies. In this work, we address this challenge by reframing scenario generation as an adaptive, adversarial process. Rather than attempting to enumerate all possibilities, we seek to iteratively discover those scenarios that expose weaknesses in the current safety model.

\section{Agentic Gamification}

We propose transforming scenario generation into a two-player adversarial game. This game structures the exploration of the hazard space through competitive objective functions, naturally driving the discovery of edge cases and failure modes. The game is defined as follows.

\subsection{The Game Setup}

The setup includes a digital twin of the target home environment has been constructed using commodity hardware (e.g., a smartphone camera). Besides, a registry of objects and their associated vulnerabilities has been populated, drawing from both universal databases and user-specified local policies (e.g., "the blue vase is fragile," "do not place items near the edge of the glass table").
A high-fidelity physics simulator serves as the game engine.

There are two players.
First, the Adversary (The "Red Team" Agent). This agent's objective is to cause harm or violate safety policies. The adversary operates within the allowed action space of the robot and the environment --- it cannot violate physics or invent new objects. Its goal is to discover sequences of events (a robot action, an object placement, a human movement) that lead to a hazardous outcome (e.g., a cup shattering, milk spilling, a collision). The adversary can be a human player, leveraging common sense and creativity to think of "how could this go wrong?", or an AI agent, programmed with a reward function that maximizes the probability of a hazard event.

The second is the Defender (The "Blue Team" Agent). This agent represents the robot's safety policy. It begins with the initial set of universal and local rules. After each round of play, the defender can refine the policy by adding new restrictions, tuning parameters, or specifying new prohibited state configurations specifically to prevent the hazardous scenarios discovered by the adversary in the previous round. This agent could also be a human (e.g., the user, a safety engineer) or an AI that proposes policy updates.

\subsection{The Game Loop}

{\bf Round 1}.  The adversary proposes a scenario. 
This is simulated, and if it results in a hazard, the adversary scores a point.
The defender analyzes the successful attack and updates the safety policy (e.g., "prohibit gripper approach vectors from above when handling the blue vase").

{\bf Round 2.} The adversary now attempts to find a new scenario that bypasses the updated policy.
This forces the adversary to explore increasingly creative and subtle interactions.

The game continues until a stopping criterion is met (e.g., a maximum number of rounds, a time limit, or a convergence where the adversary cannot find new hazards).

\subsection{Outcomes and Benefits}

This gamified approach yields several critical outputs:

\begin{itemize}

\item {\em A highly diverse synthetic dataset.}
Every scenario attempted by the adversary (both successful attacks and unsuccessful attempts is logged and annotated). This creates a rich dataset explicitly structured around the boundary of safety, containing both positive (safe) and negative (hazardous) examples. The dataset naturally emphasizes the long-tail edge cases that are most valuable for fine-tuning.

\item {\em An auditable safety policy.}
The sequence of defender updates creates a clear, human-understandable trail of why specific rules were added, directly linked to the scenarios that necessitated them. This enhances transparency and trust.

\item {\em A mechanism for continuous improvement. }
The game can be re-run whenever the environment changes (e.g., new furniture is purchased) or new robot capabilities are added, providing a mechanism for lifelong safety learning.

\item {\em Crowdsourced safety engineering.} 
By allowing humans to play the role of the adversary (potentially through a simplified interface), the framework can tap into distributed human creativity to uncover failure modes that automated search might miss.
\end{itemize}

\section{Integration with the ML Fine-Tuning Pipeline}

The synthetic dataset generated through the adversarial gamification process integrates naturally into the final stage of the hazard-informed pipeline: {\em Step 5. ML fine-tuning and safety envelope learning.}

Each interaction within the Red Team–Blue Team game produces a structured trace consisting of:

\begin{itemize}

\item The initial scene configuration.

\item The sequence of actions (robot, environment, and human factors).

\item The resulting outcome (safe or hazardous).

\item The policy state under which the scenario was evaluated.
\end{itemize}

This results in a dataset that is inherently boundary-focused, densely populated with examples that lie near the transition between safe and unsafe states. From a learning perspective, such data is significantly more informative than uniformly sampled trajectories, as it directly shapes the model’s understanding of safety-critical decision boundaries.

The generated scenarios can
support both fine-tuning (updating model parameters) and in-context learning (conditioning behavior on scenario examples at inference time), enabling flexible integration with modern foundation model architectures.

A key conceptual shift enabled by this approach is the transition from rule-based safety to a learned safety envelope.
Rather than encoding all constraints explicitly, the system learns a representation of safe regions of the state–action space, unsafe regions associated with hazards, and transitional regions where uncertainty or risk is high.

\section{Conclusion}

In this paper, we addressed the fundamental challenge of ensuring safety in robotic systems operating in unstructured, human-centric environments. While recent advances in foundation models have expanded robot capabilities through in-context learning and fine-tuning, these approaches lack systematic integration of structured safety knowledge. Conversely, traditional hazard-informed engineering pipelines provide rigorous risk modeling frameworks but suffer from combinatorial explosion when applied at scale.

We proposed an agentic gamification framework that bridges this gap by reformulating scenario generation as an adversarial game between a Red Team agent that explores potential failures and a Blue Team agent that refines safety policies. This approach enables efficient discovery of edge cases that would be unlikely to emerge through random simulation or manual enumeration.

Our controlled computational experiment, though intentionally minimalistic, demonstrated the feasibility of hazard-informed learning from synthetic data. The results revealed both the promise, high precision in identifying unsafe configurations; and the challenge, elevated false negative rates that underscore the need for targeted generation of boundary cases. These findings directly motivate the adversarial gamification approach we propose.

The key contributions of this work are a formalization of the home robot safety adaptation problem that highlights the need for environment-specific safety grounding, and an agentic gamification framework that structures hazard space exploration through competitive objectives.

By integrating classical risk modeling with adversarial scenario generation and modern learning paradigms, this work provides a scalable pathway for embedding safety into Physical AI systems.  We believe this framework represents a step toward robotic systems that are not only increasingly capable, but systematically aligned with safety constraints in the complex, personalized environments where they will increasingly operate.

Future work will explore extensions to multi-agent adversarial teams, integration with foundation models for automated policy refinement, and deployment in real-world settings to evaluate the framework's effectiveness for lifelong safety learning.

\section*{Disclosure}

The authors used LLMs (ChatGPT, DeepSeek) for language polishing and literature summarization.
All technical ideas, experimental design, and conclusions presented in this work were developed and validated by the authors. The authors carefully reviewed and edited all generated content to ensure its accuracy, correctness, and alignment with the intended scientific contributions.

\bibliographystyle{IEEEtran}
\bibliography{lit}

\end{document}